\documentclass{article}

\usepackage{microtype}
\usepackage{graphicx}
\usepackage{booktabs} 

\usepackage{hyperref}

\usepackage[accepted]{icml2019}

\usepackage{wrapfig}
\usepackage{amsmath,amsfonts,bm}
\usepackage[color=yellow,textwidth=\marginparwidth]{todonotes}
\usepackage{caption}
\usepackage{subcaption}
\usepackage{tabularx}
\usepackage{lipsum}
\usepackage{color}
\usepackage{xr-hyper}
\usepackage[subtle,title=normal,sections=normal,margins=normal,indent=normal,bibliography=normal]{savetrees} %

\newif\ifcomments
\commentsfalse
\commentstrue  
\ifcomments
\newcommand{\cg}[1]{\textcolor{purple}{\bf\small [#1 --CG]}}
\newcommand{\nando}[1]{\textcolor{blue}{\bf\small [#1 --N]}}

\newcommand{\pedro}[1]{\textcolor{orange}{\bf\small [#1]}}

\else
\newcommand{\cg}[1]{}
\newcommand{\nando}[1]{}
\newcommand{\pedro}[1]{}
\fi

\makeatletter
\newcommand*{\addFileDependency}[1]{
  \typeout{(#1)}
  \@addtofilelist{#1}
  \IfFileExists{#1}{}{\typeout{No file #1.}}
}
\makeatother
\newcommand*{\myexternaldocument}[1]{%
    \externaldocument{#1}%
    \addFileDependency{#1.tex}%
    \addFileDependency{#1.aux}%
}
\myexternaldocument{supplementary}

\icmltitlerunning{Social Influence as Intrinsic Motivation for Multi-Agent Deep RL}

\begin{document}

\twocolumn[
\icmltitle{Social Influence as Intrinsic Motivation \\ for Multi-Agent Deep Reinforcement Learning}




\begin{icmlauthorlist}
\icmlauthor{Natasha Jaques}{mit,dm}
\icmlauthor{Angeliki Lazaridou}{dm}
\icmlauthor{Edward Hughes}{dm}
\icmlauthor{Caglar Gulcehre}{dm}
\icmlauthor{Pedro A. Ortega}{dm}
\icmlauthor{DJ Strouse}{princeton}
\icmlauthor{Joel Z. Leibo}{dm}
\icmlauthor{Nando de Freitas}{dm}
\end{icmlauthorlist}

\icmlaffiliation{mit}{Media Lab, Massachusetts Institute of Technology, Cambridge, USA}
\icmlaffiliation{dm}{Google DeepMind, London, UK}
\icmlaffiliation{princeton}{Princeton University, Princeton, USA}

\icmlcorrespondingauthor{Natasha Jaques}{jaquesn@mit.edu}
\icmlcorrespondingauthor{Angeliki Lazaridou}{angeliki@google.com}

\icmlkeywords{Machine Learning, Multi-Agent, Reinforcement Learning, Intrinsic Motivation, Causal Inference, Counterfactuals, Game Theory, Social Dilemma, ICML}

\vskip 0.3in
]



\printAffiliationsAndNotice{}  

\begin{abstract}
We propose a unified mechanism for achieving coordination and communication in Multi-Agent Reinforcement Learning (MARL), through rewarding agents for having causal influence over other agents' actions. Causal influence is assessed using counterfactual reasoning. At each timestep, an agent simulates alternate actions that it could have taken, and computes their effect on the behavior of other agents. Actions that lead to bigger changes in other agents' behavior are considered influential and are rewarded. We show that this is equivalent to rewarding agents for having high mutual information between their actions. 
Empirical results demonstrate that influence leads to enhanced coordination and communication in challenging social dilemma environments, dramatically increasing the learning curves of the deep RL agents, and leading to more meaningful learned communication protocols. The influence rewards for all agents can be computed in a decentralized way by enabling agents to learn a model of other agents using deep neural networks. In contrast, key previous works on emergent communication in the MARL setting were unable to learn diverse policies in a decentralized manner and had to resort to centralized training. 
Consequently, the influence reward opens up a window of new opportunities for research in this area.




\end{abstract}

\section{Introduction}

Intrinsic Motivation for Reinforcement Learning (RL) refers to reward functions that allow agents to learn useful behavior across a variety of tasks and environments, sometimes in the absence of environmental reward \citep{singh04}. Previous approaches to intrinsic motivation often focus on curiosity (e.g. \citet{pathak2017curiosity,schmidhuber2010formal}), or empowerment (e.g. \citet{klyubin2005empowerment,mohamed2015variational}). Here, we consider the problem of deriving intrinsic social motivation from other agents in multi-agent RL (MARL). Social learning is incredibly important for humans, and has been linked to our ability to achieve unprecedented progress and coordination on a massive scale \citep{henrich2015, harari2014sapiens,laland2017,van2011social,Herrmann1360}. While some previous work has investigated intrinsic social motivation for RL (e.g. \citet{sequeira2011emerging,hughes2018inequity,peysakhovich2018prosocial}), these approaches rely on hand-crafted rewards specific to the environment, or allowing agents to view the rewards obtained by other agents. Such assumptions make it impossible to achieve independent training of MARL agents across multiple environments.

Achieving coordination among agents in MARL still remains a difficult problem. Prior work in this domain (e.g., \citet{foerster2017counterfactual,foerster2016learning}), often resorts to centralized training to ensure that agents learn to coordinate. While communication among agents could help with coordination, training emergent communication protocols also remains a challenging problem; recent empirical results underscore the difficulty of learning meaningful emergent communication protocols, even when relying on centralized training (e.g., \citet{lazaridou2018emergence,cao2018emergent,foerster2016learning}).

We propose a unified method for achieving both coordination and communication in MARL by giving agents an intrinsic reward for having a causal influence on other agents' actions. Causal influence is assessed using counterfactual reasoning; at each timestep, an agent simulates alternate, counterfactual actions that it could have taken, and assesses their effect on another agent's behavior. Actions that lead to relatively higher change in the other agent's behavior are considered to be highly influential and are rewarded. We show how this reward is related to maximizing the mutual information between agents' actions, and hypothesize that this inductive bias will drive agents to learn coordinated behavior. Maximizing mutual information as a form of intrinsic motivation has been studied in the literature on empowerment (e.g. \citet{klyubin2005empowerment,mohamed2015variational}). Social influence can be seen as a novel, social form of empowerment. 

To study our influence reward, we adopt the Sequential Social Dilemma (SSD) multi-agent environments of \citet{leibo2017multi}. Through a series of three experiments, we show that the proposed social influence reward allows agents to learn to coordinate and communicate more effectively in these SSDs. We train recurrent neural network policies directly from pixels, and show in the first experiment that deep RL agents trained with the proposed social influence reward learn effectively and attain higher collective reward than powerful baseline deep RL agents, which often completely fail to learn.


In the second experiment, the influence reward is used to directly train agents to use an explicit communication channel. We demonstrate that the communication protocols trained with the influence reward are more meaningful and effective for obtaining better collective outcomes. Further, we find a significant correlation between being influenced through communication messages and obtaining higher individual reward, suggesting that influential communication 
is beneficial to the agents that receive it. By examining the learning curves in this second experiment, we again find that the influence reward is essential to allow agents to learn to coordinate.

Finally, we show that influence agents can be trained independently, when each agent is equipped with an internal neural network \textit{Model of Other Agents} (MOA), which has been trained to predict the actions of every other agent. The agent can then simulate counterfactual actions and use its own internal MOA to predict how these will affect other agents, thereby computing its own intrinsic influence reward. Influence agents can thus learn socially, only through observing other agents' actions, and without requiring a centralized controller or access to another agent's reward function. 
Therefore, the influence reward offers us a simple, general and effective way of overcoming long-standing unrealistic assumptions and limitations in this field of research, including centralized training and the sharing of reward functions or policy parameters. Moreover, both the influence rewards as well as the agents' policies can be learned directly from pixels using expressive deep recurrent neural networks. In this third experiment, the learning curves once again show that the influence reward is essential for learning to coordinate in these complex domains.


The paper is structured as follows. We describe the environments in Section \ref{sec:ssd}, and the MARL setting in Section~\ref{sec:marl}. Section~\ref{sec:si_reward} introduces the basic formulation of the influence reward, Section~\ref{sec:comm_model} extends it with the inclusion of explicit communication protocols, and Section~\ref{sec:tom} advances it by including models of other agents to achieve independent training. Each of these three sections presents experiments and results that empirically demonstrate the efficacy of the social influence reward. Related work is presented in Section~\ref{sec:related_work}. Finally, more details about the causal inference procedure are given in Section~\ref{sec:causal_details}.

\section{Sequential Social Dilemmas}
\label{sec:ssd}
Sequential Social Dilemmas (SSDs) \citep{leibo2017multi} are partially observable, spatially and temporally extended multi-agent games with a game-theoretic payoff structure. An individual agent can obtain higher reward in the short-term by engaging in defecting, non-cooperative behavior (and thus is greedily motivated to defect), but the total payoff per agent will be higher if all agents cooperate. Thus, the collective reward obtained by a group of agents in these SSDs gives a clear signal about how well the agents learned to cooperate \citep{hughes2018inequity}.

We experiment with two SSDs in this work, a public goods game \textit{Cleanup}, and a public pool resource game \textit{Harvest}. In both games apples (green tiles) provide the rewards, but are a limited resource. Agents must coordinate harvesting apples with the behavior of other agents in order to achieve cooperation (for further details see Section \ref{sec:ssd} of the Supplementary Material). For reproducibility, the code for these games has been made available in open-source.\footnote{\url{https://github.com/eugenevinitsky/sequential\_social\_dilemma\_games}}

As the Schelling diagrams in Figure \ref{fig:schelling} of the Supplementary Material reveal, all agents would benefit from learning to cooperate in these games, because even agents that are being exploited get higher reward than in the regime where more agents defect. However, traditional RL agents struggle to learn to coordinate or cooperate to solve these tasks effectively \citep{hughes2018inequity}. Thus, these SSDs represent challenging benchmark tasks for the social influence reward. Not only must influence agents learn to coordinate their behavior to obtain high reward, they must also learn to cooperate. 


\section{Multi-Agent RL for SSDs}
\label{sec:marl}
We consider a MARL Markov game defined by the tuple $\langle S, T, A, r \rangle$, in which multiple agents are trained to independently maximize their own individual reward; agents do not share weights. The environment state is given by $s \in S$. At each timestep $t$, each agent $k$ chooses an action $a^k_t \in A$. The actions of all $N$ agents are combined to form a joint action $\boldsymbol{a}_t={[a^0_t, ... a^N_t]}$, which produces a transition in the environment $T(s_{t+1}|\boldsymbol{a}_t,s_t)$, according to the state transition distribution $T$. Each agent then receives its own reward $r^k(\boldsymbol{a}_t, s_t)$, which may depend on the actions of other agents. A history of these variables over time is termed a trajectory, $\tau=\left\{ s_{t}, \boldsymbol{a}_t, \boldsymbol{r}_t\right\} _{t=0}^{T}$. We consider a partially observable setting in which the $k$th agent can only view a portion of the true state, $s^k_t$. Each agent seeks to maximize its own total expected discounted future reward, $R^k = \sum_{i=0}^\infty \gamma^i r^k_{t+i}\,$, where $\gamma$ is the discount factor. A distributed asynchronous advantage actor-critic (A3C) approach \citep{mnih2016asynchronous} is used to train each agent's policy $\pi^k$. 


Our neural networks consist of a convolutional layer, fully connected layers, a Long Short Term Memory (LSTM) recurrent layer \citep{gers1999learning}, and linear layers. All networks take images as input and output both the policy $\pi^k$ and the value function $V^{\pi_k}(s)$, but some network variants consume additional inputs and output either communication policies or models of other agents' behavior. We will refer to the internal LSTM state of the $k$th agent at timestep $t$ as $u^k_t$.

\section{Basic Social Influence}
\label{sec:si_reward}
Social influence intrinsic motivation gives an agent additional reward for having a causal influence on another agent's actions. Specifically, it modifies an agent's immediate reward so that it becomes $r^k_t = \alpha e^k_t + \beta c^k_t$, where $e^k_t$ is the extrinsic or environmental reward, and $c^k_t$ is the causal influence reward.

To compute the causal influence of one agent on another, suppose there are two agents, $k$ and $j$, and that agent $j$ is able to condition its policy on agent $k$'s action at time $t$, $a^k_t$. Thus, agent $j$ computes the probability of its next action as $p(a^j_{t}| a^k_t, s^j_t)$. We can then intervene on $a^k_t$ by replacing it with a counterfactual action, $\tilde{a}^k_t$. This counterfactual action is used to compute a new distribution over $j$'s next action, $p(a^j_{t} | \tilde{a}^k_t, s^j_t)$. Essentially, agent $k$ asks a retrospective question: ``How would $j$'s action change if I had acted differently in this situation?''.

By sampling several counterfactual actions, and averaging the resulting policy distribution of $j$ in each case, we obtain the marginal policy of $j$, $p(a^j_{t}|s^j_t) = \sum_{\tilde{a}^k_t}p(a^j_{t}|\tilde{a}^k_t,s^j_t)p(\tilde{a}^k_t|s^j_t)$ ---in other words, $j$'s policy if it did not consider agent $k$. 
The discrepancy between the marginal policy of $j$ and the conditional policy of $j$ given $k$'s action is a measure of the causal influence of $k$ on $j$; it gives the degree to which $j$ changes its planned action distribution because of $k$'s action. Thus, the causal influence reward for agent $k$ is:
\begin{align}
\vspace{-0.1cm}
c^k_t &= \sum_{j=0, j \neq k}^N \Bigl[D_{KL}[p(a^j_{t} \mid a^k_t, s^j_t) \Bigl\| \sum_{\tilde{a}^k_t}p(a^j_{t} \mid \tilde{a}^k_t, s^j_t)p(\tilde{a}^k_t \mid s^j_t) ] \Bigr] \nonumber \\
&= \sum_{j=0, j \neq k}^N \Bigl[D_{KL}[p(a^j_{t} \mid a^k_t, s^j_t) \Bigl\| p(a^j_{t} \mid s^j_t)] \Bigr]\, .\label{eq:ci_reward}
\end{align}
Note that it is possible to use a divergence metric other than KL; we have found empirically that the influence reward is robust to the choice of metric.

The reward in Eq. \ref{eq:ci_reward} is related to the mutual information (MI) between the actions of agents $k$ and $j$, $I(a^k;a^j|s)$. As the reward is computed over many trajectories sampled independently from the environment, we obtain a Monte-Carlo estimate of $I(a^k;a^j|s)$. In expectation, the influence reward incentivizes agents to maximize the mutual information between their actions. The proof is given in Section \ref{sec:mi_connection} of the Supplementary Material. Intuitively, training agents to maximize the MI between their actions results in more coordinated behavior.

Moreover, the variance of policy gradient updates increases as the number of agents in the environment grows \citep{lowe2017multi}. This issue can hinder convergence to equilibrium for large-scale MARL tasks. Social influence can reduce the variance of policy gradients by introducing explicit dependencies across the actions of each agent. This is because the conditional variance of the gradients an agent is receiving will be less than or equal to the marginalized variance. 

Note that for the basic influence model we make two assumptions: 1) we use centralized training to compute $c^k_t$ directly from the policy of agent $j$, and 2) we assume that influence is unidirectional: agents trained with the influence reward can only influence agents that are not trained with the influence reward (the sets of influencers and influencees are disjoint, and the number of influencers is in $[1,N-1]$). Both of these assumptions are relaxed in later sections. Further details, as well as further explanation of the causal inference procedure (including causal diagrams) are available in Section \ref{sec:causal_details}.

\subsection{Experiment I: Basic Influence}
\begin{figure}[h]
\centering
\begin{subfigure}{.49\linewidth}
  \centering
  \includegraphics[width=\linewidth]{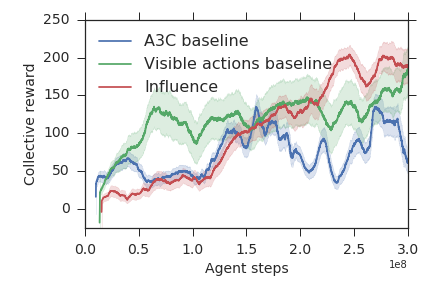}
  \caption{\textit{Cleanup}}
  \label{fig:results_cc_huangpu}
\end{subfigure}%
\begin{subfigure}{.49\linewidth}
  \centering
  \includegraphics[width=\linewidth]{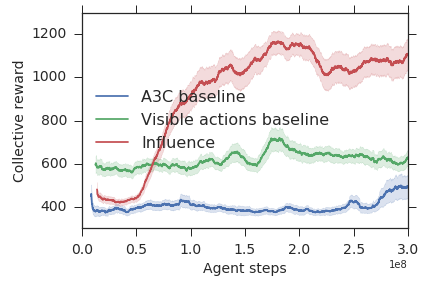}
  \caption{\textit{Harvest}}
  \label{fig:results_cc_tragedy}
\end{subfigure}
\caption{Total collective reward obtained in Experiment 1. Agents trained with influence (red) significantly outperform the baseline and ablated agents. In Harvest, the influence reward is essential to achieve any meaningful learning.}
\label{fig:cc_results}
\end{figure}

Figure \ref{fig:cc_results} shows the results of testing agents trained with the basic influence reward against standard A3C agents, and an ablated version of the model in which agents do not receive the influence reward, but are able to condition their policy on the actions of other agents (even when the other agents are not within the agent's partially observed view of the environment). We term this ablated model the \textit{visible actions baseline}. In this and all other results figures, we measure the total collective reward obtained using the best hyperparameter setting tested with 5 random seeds each. Error bars show a 99.5\% confidence interval (CI) over the random seeds, computed within a sliding window of 200 agent steps. We use a curriculum learning approach which gradually increases the weight of the social influence reward over $C$ steps ($C\in[0.2-3.5]\times 10^8$); this sometimes leads to a slight delay before the influence models' performance improves. 

As is evident in Figures \ref{fig:results_cc_huangpu} and \ref{fig:results_cc_tragedy}, introducing an awareness of other agents' actions helps, but having the social influence reward eventually leads to significantly higher collective reward in both games. Due to the structure of the SSD games, we can infer that agents that obtain higher collective reward learned to cooperate more effectively. 
In the \textit{Harvest} MARL setting, it is clear that the influence reward is essential to achieve any reasonable learning.

To understand how social influence helps agents achieve cooperative behavior, we investigated the trajectories produced by high scoring models in both \textit{Cleanup} and \textit{Harvest}; the analysis revealed interesting behavior. As an example, in the \textit{Cleanup} video available here: \url{https://youtu.be/iH_V5WKQxmo}
a single agent (shown in purple) was trained with the social influence reward. Unlike the other agents, which continue to move and explore randomly while waiting for apples to spawn, the influencer only traverses the map when it is pursuing an apple, then stops. 
The rest of the time it stays still. 

\begin{wrapfigure}{l}{0.4\linewidth}
\includegraphics[width=\linewidth]{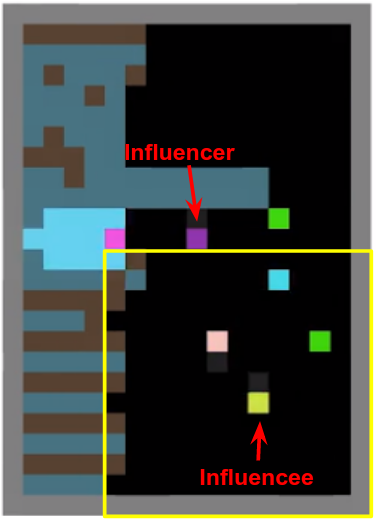}
\caption{A moment of high influence when the purple influencer signals the presence of an apple (green tiles) outside the yellow influencee's field-of-view (yellow outlined box).}
\label{fig:influence_moment1}
\end{wrapfigure}
Figure \ref{fig:influence_moment1} shows a moment of high influence between the influencer and the yellow influencee. The influencer has chosen to move towards an apple that is outside of the ego-centric field-of-view of the yellow agent. Because the influencer only moves when apples are available, this signals to the yellow agent that an apple must be present above it which it cannot see. This changes the yellow agent's distribution over its planned action, $p(a^j_t|a^k_t,s^j_t)$, and allows the purple agent to gain influence. A similar moment occurs when the influencer signals to an agent that has been cleaning the river that no apples have appeared by staying still (see Figure \ref{fig:influence_moment2} in the Supplementary Material).

In this case study, the influencer agent learned to use its own actions as a binary code which signals the presence or absence of apples in the environment. We observe a similar effect in \textit{Harvest}. This type of action-based communication could be likened to the bee waggle dance discovered by \citet{vonFrisch1969}. Evidently, the influence reward gave rise not only to cooperative behavior, but to emergent communication. 

It is important to consider the limitations of the influence reward. Whether it will always give rise to cooperative behavior may depend on the specifics of the environment and task, and tuning the trade-off between environmental and influence reward. 
Although influence is arguably necessary for coordination (e.g. two agents coordinating to manipulate an object must have a high degree of influence between their actions), it may be possible to influence another agent in a non-cooperative way. The results provided here show that the influence reward did lead to increased cooperation, in spite of cooperation being difficult to achieve in these environments.


\section{Influential Communication}
\label{sec:comm_model}
Given the above results, 
we next experiment with using the influence reward to train agents to use an explicit communication channel. We take some inspiration from research drawing a connection between influence and communication in human learning. According to \citet{melis2010human}, human children rapidly learn to use communication to influence the behavior of others when engaging in cooperative activities. They explain that ``this ability to influence the partner via communication has been interpreted as evidence for a capacity to form shared goals with others'', and that this capacity may be ``what allows humans to engage in a wide range of cooperative activities''. 

Thus, we equip agents with an explicit communication channel, similar to the approach used by \citet{foerster2016learning}. 
At each timestep, each agent $k$ chooses a discrete communication symbol $m^k_t$; these symbols are concatenated into a combined message vector $\boldsymbol{m}_t = {[m^0_t, m^1_t ... m^N_t]}$, for $N$ agents. This message vector $\boldsymbol{m}_t$ is then given as input to every other agent in the next timestep. Note that previous work has shown that self-interested agents do not learn to use this type of ungrounded, \textit{cheap talk} communication channel effectively \citep{Crawford:Sobel:1982,cao2018emergent,foerster2016learning,lazaridou2018emergence}.

\begin{figure}[h]
\vspace{-0.2cm}
\centering
\includegraphics[width=.9\linewidth]{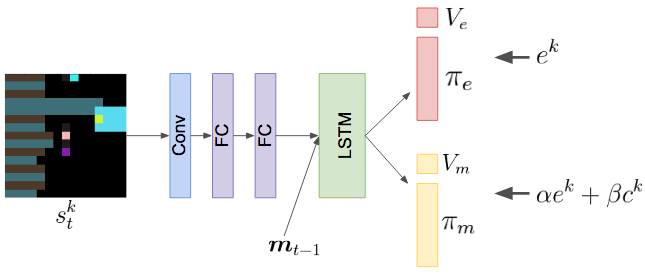}
\caption{The communication model has two heads, which learn the environment policy, $\pi_e$, and a policy for emitting communication symbols, $\pi_m$. Other agents' communication messages $\boldsymbol{m}_{t-1}$ are input to the LSTM.}
\label{fig:comm_model}
\vspace{-0.2cm}
\end{figure}

To train the agents to communicate, we augment our initial network with an additional A3C output head, that learns a communication policy $\pi_m$ and value function $V_m$ to determine which symbol to emit (see Figure \ref{fig:comm_model}). The normal policy and value function used for acting in the environment, $\pi_e$ and $V_e$, are trained only with environmental reward $e$. We use the influence reward as an additional incentive for training the communication policy, $\pi_m$, such that $r= \alpha e + \beta c$. Counterfactuals are employed to assess how much influence an agent's communication message from the previous timestep, $m^k_{t-1}$, has on another agent's action, $a^j_t$, where: 
\begin{align}
    c^k_t &= \sum_{j=0, j \neq k}^N \Bigl[D_{KL}[p(a^j_t \mid m^k_{t-1}, s^j_t) \Bigl\| p(a^j_t \mid s^j_t)]\Bigr]
\end{align}

Importantly, rewarding influence through a communication channel does not suffer from the limitation mentioned in the previous section, i.e. that it may be possible to influence another agent in a non-cooperative way. We can see this for two reasons. First, there is nothing that compels agent $j$ to act based on agent $k$'s communication message; if $m^k_t$ does not contain valuable information, $j$ is free to ignore it. 
Second, because $j$'s action policy $\pi_e$ is trained only with environmental reward, $j$ will only change its intended action as a result of observing $m^k_t$ (i.e. be influenced by $m^k_t$) if it contains information that helps $j$ to obtain environmental reward. Therefore, we hypothesize that influential communication must provide useful information to the listener.

\subsection{Experiment II: Influential Communication}

\begin{figure}[h]
\centering
\begin{subfigure}{.49\linewidth}
  \centering
  \includegraphics[width=\linewidth]{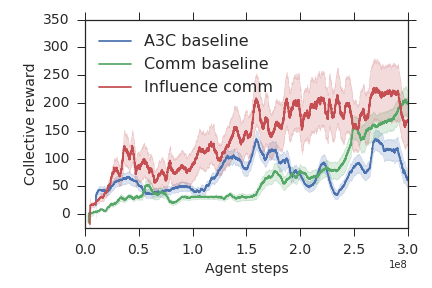}
  \caption{\textit{Cleanup}}
  \label{fig:results_comm_huangpu}
\end{subfigure}%
\begin{subfigure}{.49\linewidth}
  \centering
  \includegraphics[width=\linewidth]{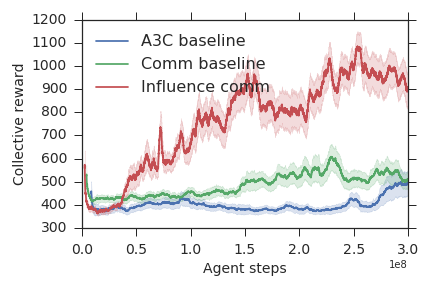}
  \caption{\textit{Harvest}}
  \label{fig:results_comm_tragedy}
\end{subfigure}
\caption{Total collective reward for deep RL agents with communication channels. Once again, the influence reward is essential to improve or achieve any learning.}
\label{fig:results_comm}
\end{figure}

Figure \ref{fig:results_comm} shows the collective reward obtained when training the agents to use an explicit communication channel. Here, the ablated model has the same structure as in Figure \ref{fig:comm_model}, but the communication policy $\pi_m$ is trained only with environmental reward.
We observe that the agents incentivized to communicate via the social influence reward learn faster, and achieve significantly higher collective reward for the majority of training in both games. In fact, in the case of \textit{Cleanup}, we found that $\alpha=0$ in the optimal hyperparameter setting, meaning that it was most effective to train the communication head with zero extrinsic reward (see Table \ref{tab:hparams_comm} in the Supplementary Material). This suggests that influence alone can be a sufficient mechanism for training an effective communication policy. In \textit{Harvest}, once again influence  is critical to allow agents to learn coordinated policies and attain high reward.

To analyze the communication behaviour learned by the agents, we introduce three metrics, partially inspired by \cite{bogin2018emergence}. \textit{Speaker consistency}, is a normalized score $\in [0,1]$ which assesses the entropy of $p(a^k|m^k)$ and $p(m^k|a^k)$  to determine how consistently a \textit{speaker} agent emits a particular symbol when it takes a particular action, and vice versa (the formula is given in the Supplementary Material Section \ref{app:comm}). We expect this measure to be high if, for example, the speaker
always emits the same symbol when it is cleaning the river. We also introduce two measures of \textit{instantaneous coordination} (IC), which are both measures of mutual information (MI): (1) symbol/action IC $=I(m^k_t;\,a^j_{t+1})$ measures the MI between the influencer/speaker's symbol and the influencee/listener's next action, and (2) action/action IC $=I(a^k_t;\,a^j_{t+1})$ measures the MI between the influencer's action and the influencee's next action. To compute these measures we first average over all trajectory steps, then take the maximum value between any two agents, to determine if any pair of agents are coordinating. 
Note that these measures are all \textit{instantaneous}, as they consider only short-term dependencies across two consecutive timesteps, and cannot capture if an agent communicates influential compositional messages, i.e. information that requires several consecutive symbols to transmit and only then affects the other agents behavior.

\begin{figure}[h]
\includegraphics[width=\linewidth]{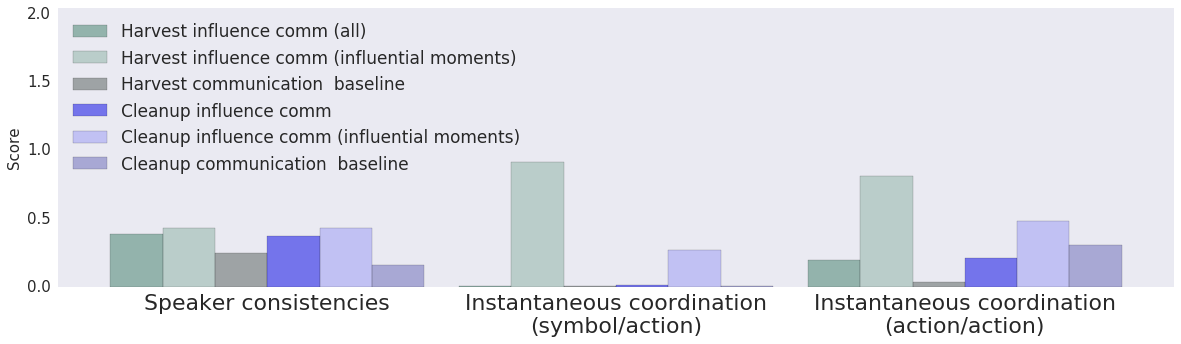} 
\caption{Metrics describing the quality of learned communication protocols. The models trained with influence reward exhibit more consistent communication and more coordination, especially in moments where influence is high.}
\label{fig:comm_analysis}
\vspace{-0.2cm}
\end{figure}
Figure \ref{fig:comm_analysis} presents the results. The speaker consistencies metric reveals that influence agents more unambiguously communicate about their own actions than baseline agents, indicating that the emergent communication is more meaningful. The IC metrics demonstrate 
that baseline agents show almost no signs of co-ordinating behavior with communication, i.e. speakers saying A and listeners doing B consistently. This result is aligned with both theoretical results in cheap-talk literature \citep{Crawford:Sobel:1982}, and recent empirical results in MARL (e.g. \citet{,foerster2016learning,lazaridou2018emergence,cao2018emergent}). 

In contrast, we do see high IC between influence agents, but only when we limit the analysis to timesteps on which influence was greater than or equal to the mean influence (cf. \textit{influential moments} in Figure~\ref{fig:comm_analysis}). Inspecting the results reveals a common pattern: influence is sparse in time. An agent's influence is only greater than its mean influence in less than 10\% of timesteps. Because the listener agent is not compelled to listen to any given speaker, listeners selectively listen to a speaker only when it is beneficial, and influence cannot occur all the time. Only when the listener decides to change its action based on the speaker's message does influence occur, and in these moments we observe high $I(m^k_t;a^j_{t+1})$. It appears the influencers have learned a strategy of communicating meaningful information about their own actions, and gaining influence when this becomes relevant enough for the listener to act on it. 

Examining the relationship between the degree to which agents were influenced by communication and the reward they obtained gives a compelling result: agents that are the most influenced also achieve higher individual environmental reward. We sampled 100 different experimental conditions (i.e., hyper-parameters and random seeds) for both games, and normalized and correlated the influence and individual rewards. We found that agents who are more often influenced tend to achieve higher task reward in both \textit{Cleanup}, $\rho=.67$, $p\textless 0.001$, and \textit{Harvest}, $\rho=.34$, $p\textless 0.001$. This supports the hypothesis that in order to influence another agent via communication, the communication message should contain information that helps the listener maximize its own environmental reward. Since better listeners/influencees are more successful in terms of task reward, we have evidence that useful information was transmitted to them. 

This result is promising, but may depend on the specific experimental approach taken here, in which agents interact with each other repeatedly. In this case, there is no advantage to the speaker for communicating unreliable information (i.e. lying), because it would lose influence with the listener over time. This may not be guaranteed in one-shot interactions. However, given repeated interactions, the above results provide empirical evidence that social influence as intrinsic motivation allows agents to learn meaningful communication protocols when this is otherwise not possible. 

\section{Modeling Other Agents}
\label{sec:tom}

\begin{figure}[h]
\includegraphics[width=\linewidth]{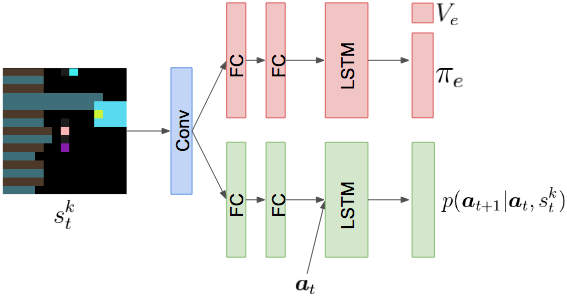}
\caption{The Model of Other Agents (MOA) architecture learns both an RL policy $\pi_e$, and a supervised model that predicts the actions of other agents, $\boldsymbol{a}_{t+1}$. The supervised model is used for internally computing the influence reward.}
\label{fig:tom_model}
\vspace{-0.4cm}
\end{figure}

Computing the causal influence reward as introduced in Section \ref{sec:si_reward} requires knowing the probability of another agent's action given a counterfactual,  
which we previously solved by using a centralized training approach in which agents could access other agents' policy networks. While using a centralized training framework is common in MARL (e.g. \citet{foerster2017counterfactual, foerster2016learning}), it is less realistic than a scenario in which each agent is trained independently. We can relax this assumption and achieve independent training by equipping each agent with its own internal \textit{Model of Other Agents} (MOA).
The MOA consists of a second set of fully-connected and LSTM layers connected to the agent's convolutional layer (see Figure \ref{fig:tom_model}), and is trained to predict all other agents' next actions given their previous actions, and the agent's egocentric view of the state: $p(\boldsymbol{a}_{t+1}|\boldsymbol{a}_t, s^k_t)$. The MOA is trained using observed action trajectories and cross-entropy loss. 

A trained MOA can be used to compute the social influence reward in the following way. Each agent can ``imagine" counterfactual actions that it could have taken at each timestep, and use its internal MOA to predict the effect on other agents. It can then give itself reward for taking actions that it estimates were the most influential. This has an intuitive appeal, because it resembles how humans reason about their effect on others \citep{ferguson2010}. We often find ourselves asking counterfactual questions of the form, ``How would she have acted if I had done something else in that situation?'', which we answer using our internal model of others. 

Learning a model of $p(a^j_{t+1} | a^k_t, s^k_t)$ requires implicitly modeling both other agents' internal states and behavior, as well as the environment transition function. If the model is inaccurate, this would lead to noisy estimates of the causal influence reward. To compensate for this, 
We only give the influence reward to an agent ($k$) when the agent it is attempting to influence ($j$) is within its field-of-view, because the estimates of $p(a^j_{t+1} | a^k_t, s^k_t)$ are 
more accurate when $j$ is visible to $k$.\footnote{This contrasts with our previous models in which the influence reward was obtained even from non-visible agents.} 
This constraint could have the side-effect of encouraging agents to stay in closer proximity. However, an intrinsic social reward encouraging proximity is reasonable given that humans seek affiliation and to spend time near other people \citep{tomasello2009we}. 

\subsection{Experiment III: Modeling Other Agents}
\begin{figure}[h]
\vspace{-0.2cm}
\centering
\begin{subfigure}{.49\linewidth}
  \centering
  \includegraphics[width=\linewidth]{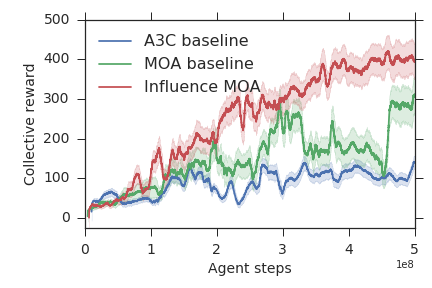}
  \caption{\textit{Cleanup}}
  \label{fig:results_tom_huangpu}
\end{subfigure}%
\begin{subfigure}{.49\linewidth}
  \centering
  \includegraphics[width=\linewidth]{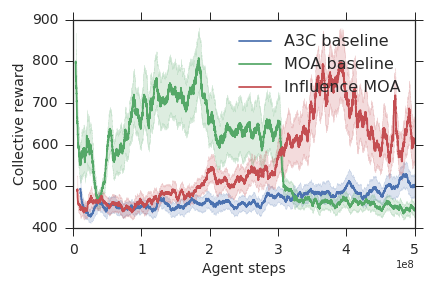}
  \caption{\textit{Harvest}}
  \label{fig:results_tom_tragedy}
\end{subfigure}
\caption{Total collective reward for MOA models. Again, intrinsic influence consistently improves learning, with the powerful A3C agent baselines not being able to learn.}
\label{fig:results_moa}
\vspace{-0.3cm}
\end{figure}

As before, we allow the policy LSTM of each agent to condition on the actions of other agents in the last timestep (actions are visible). We compare against an ablated version of the architecture shown in Figure \ref{fig:tom_model}, which does not use the output of the MOA to compute a reward; rather, the MOA can be thought of as an unsupervised auxiliary task that may help the model to learn a better shared embedding layer, encouraging it to encode information relevant to predicting other agents' behavior. Figure \ref{fig:results_moa} shows the collective reward obtained for agents trained with a MOA module. While we see that the auxiliary task does help to improve reward over the A3C baseline, the influence agent gets consistently higher collective reward. These results demonstrate that the influence reward can be effectively computed using an internal MOA, and thus agents can learn socially but independently, optimizing for a social reward without a centralized controller. 

Agents with influence achieve higher collective reward than the previous state-of-the-art for these environments ($275$ for \textit{Cleanup} and $750$ for \textit{Harvest}) \citep{hughes2018inequity}. This is compelling, given that previous work relied on the assumption that agents could view one another's rewards; we make no such assumption, instead relying only on agents viewing each other's actions. Table \ref{tab:final_scores} of the Supplementary Material  gives the final collective reward obtained in previous work, and by each influence model for all three experiments.

\section{Related work}
\label{sec:related_work}
Several attempts have been made to develop intrinsic social rewards.\footnote{Note that \textit{intrinsic} is not a synonym of \textit{internal}; other people can be intrinsically motivating \citep{stavropoulos2013research}.} \citet{sequeira2011emerging} developed hand-crafted rewards for a foraging environment, in which agents were punished for eating more than their fair share of food. Another approach 
gave agents an emotional intrinsic reward based on their perception of their neighbours' cooperativeness in a networked version of the iterated prisoner's dilemma, but is limited to scenarios in which it is possible to directly classify each action as cooperative or non-cooperative \citep{yu2013emotional}. This is untenable in complex settings with long-term strategies, such as the SSDs under investigation here. 

Some approaches allow agents to view each others' rewards in order to optimize for collective reward. \citet{peysakhovich2018prosocial} show that if even a single agent is trained to optimize for others' rewards, it can significantly help the group. \citet{hughes2018inequity} introduced an inequity aversion motivation, which penalized agents if their rewards differed too much from those of the group. 
\citet{liu2014optimal} train agents to learn their own optimal reward function in a cooperative, multi-agent setting with known group reward. 
However, the assumption that agents can view and optimize for each others' rewards may be unrealistic. Thus, recent work explores training agents that learn when to cooperate based solely on their own past rewards \citep{peysakhovich2017consequentialist}.

Training agents to learn emergent communication protocols has been explored \citep{foerster2016learning, cao2018emergent, choi2018compositional, lazaridou2018emergence, bogin2018emergence}, with many authors finding that selfish agents do not learn to use an ungrounded, \textit{cheap talk} communication channel effectively. \citet{Crawford:Sobel:1982} find that in theory, the information communicated is proportional to the amount of common interest; thus, as agents' interests diverge, no communication is to be expected. And while communication can emerge when agents are prosocial \citep{foerster2016learning,lazaridou2018emergence}, curious \citep{oudeyer2006discovering, oudeyer2016evolution, forestier2017unified}, or hand-crafted \citep{crandall17}, self-interested agents do not to learn to communicate \citep{cao2018emergent}. We have shown that the social influence reward can encourage agents to learn to communicate more effectively in complex environments.

Our MOA is related to work on machine theory of mind \citep{rabinowitz2018machine}, which demonstrated that a model trained to predict agents' actions can model false beliefs. LOLA agents model the impact of their policy on the parameter updates of other agents, and directly incorporate this into the agent's own learning rule \citep{foerster2018learning}.

\citet{barton2018measuring} propose causal influence as a way to measure coordination between agents, specifically using Convergence Cross Mapping (CCM) to analyze the degree of dependence between two agents' policies. The limitation if CCM is that estimates of causality are known to degrade in the presence of stochastic effects \citep{tajima2015untangling}. Counterfactual reasoning has also been used in a multi-agent setting, to marginalize out the effect of one agent on a predicted global value function estimating collective reward, and thus obtain an improved baseline for computing each agent's advantage function \citep{foerster2017counterfactual}. A similar paper shows that counterfactuals can be used with potential-based reward shaping to improve credit assignment for training a joint policy in multi-agent RL \citep{devlin2014potential}. However, once again these approaches rely on a centralized controller.

Mutual information (MI) has been explored as a tool for designing social rewards. \citet{strouse2018learning} train agents to optimize the MI between their actions and a categorical goal, as a way to signal or hide the agent's intentions. 
However, this approach depends on agents pursuing a known, categorical goal. 
\citet{guckelsberger2018new}, in pursuit of the ultimate video game adversary, develop an agent that maximizes its empowerment, minimizes the player's empowerment, and maximizes its empowerment over the player's next state. This third goal, termed {\em transfer empowerment}, is obtained by maximizing the MI between the agent's actions and the player's future state. While a social form of empowerment, the authors find that agents trained with transfer empowerment simply tend to stay near the player. Further, the agents are not trained with RL, but rather analytically compute these measures in simple grid-world environments. As such, the agent cannot learn to model other agents or the environment. 

Given the social influence reward incentivizes maximizing the mutual information between agents' actions, our work also has ties to the literature on empowerment, in which agents maximize the mutual information between their actions and their future state \citep{klyubin2005empowerment,mohamed2015variational}. Thus, our proposed reward can be seen as a novel social form of empowerment. 

\section{Details on Causal Inference}
\label{sec:causal_details}
\begin{figure}[h]
\vspace{-0.0cm}
\centering
\begin{subfigure}{.23\linewidth}
  \centering
  \includegraphics[width=\linewidth]{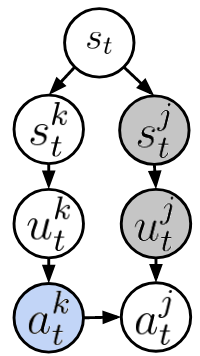}
  \caption{Basic}
  \label{fig:causal_basic}
\end{subfigure}%
\begin{subfigure}{.6\linewidth}
  \centering
  \includegraphics[width=\linewidth]{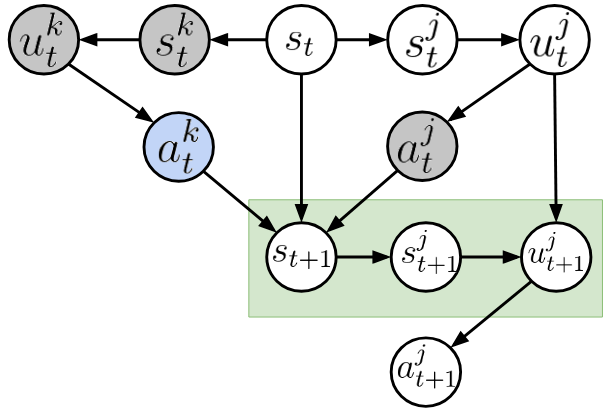}
  \caption{MOA}
  \label{fig:causal_moa}
\end{subfigure}
\caption{Causal diagrams of agent $k$'s effect on $j$'s action. Shaded nodes are conditioned on, and we intervene on $a^k_t$ (blue node) by replacing it with counterfactuals. Nodes with a green background must be modeled using the MOA module. Note that there is no backdoor path between $a^k_t$ and $s_t$ in the MOA case, since it would require traversing a collider that is not in the conditioning set.
}
\label{fig:causal}
\vspace{-0.1cm}
\end{figure}

The causal influence reward presented in Eq. \ref{eq:ci_reward} is assessed using counterfactual reasoning. Unlike a \textit{do}-calculus intervention (which estimates the general expected causal effect of one variable on another), a counterfactual involves conditioning on a set of variables observed in a given situation and asking how would the outcome have changed if some variable were different, and all other variables remained the same \citep{pearl2016causal}. This type of inquiry allows us to measure the precise causal effect of agent $k$'s action at timestep $t$, $a^k_t$, on agent $j$'s action, $a^j_t$, in the specific environment state $s_t$, providing a richer and less sparse reward for agent $k$. Computing counterfactuals requires conditioning on the correct set of observed variables to ensure there are no confounds. In our case, the conditioning set must include not only an agent's partially observed view of the environment state, $s^j_t$, but also the agent's internal LSTM state $u^j_t$, to remove any dependency on previous timesteps in the trajectory. Thus, the basic causal influence reward can be more accurately written:
\begin{align}
\vspace{-0.2cm}
c^k_t &= \sum_{j=0, j \neq k}^N \Bigl[D_{KL}[p(a^j_{t} \mid a^k_t, s^j_t, u^j_t) || p(a^j_{t} \mid s^j_t, u^j_t)]\Bigr] \, .\label{eq:ci_reward_detailed}
\end{align}
Figure \ref{fig:causal} shows the causal diagrams for computing the influence reward in both the basic case (\ref{fig:causal_basic}) and the MOA case (\ref{fig:causal_moa}). Because basic influence looks at influence between agents' actions in the same timestep, the diagram is much simpler. However, to avoid circular dependencies in the graph, it requires that agent $k$ choose its action before $j$, and therefore $k$ can influence $j$ but $j$ cannot influence $k$. If there are more than two agents, we assume a disjoint set of influencer and influencee agents, and all influencers must act first.

Computing influence across timesteps, as in the communication and MOA experiments, complicates the causal diagram, but ensures that each agent can influence every other agent. Figure \ref{fig:causal_moa} shows the diagram in the MOA case, in which we can isolate the causal effect of $a^k_t$ on $a^j_{t+1}$ because the back-door path through $s_t$ is blocked by the collider nodes at $s_{t+1}$ and $u^j_{t+1}$ \citep{pearl2016causal}. Note that it would be sufficient to condition only on $s^k_t$ in order to block all back-door paths in this case, but we show $\langle u^k_t, s^k_t, a^j_t\rangle$ as shaded because all of these are given as inputs to the MOA to help it predict $a^j_{t+1}$. For the MOA to accurately estimate $p(a^j_{t+1} | a^k_t, s^k_t)$, it must model both the environment transition function $T$, as well as aspects of the internal LSTM state of the other agent, $u^j_{t+1}$, as shown by the shaded green variables in Figure \ref{fig:causal_moa}.


This is a simple case of counterfactual reasoning, that does not require using abduction to update the probability of any unobserved variables \citep{pearl2013structural}. This is because we have built all relevant models, know all of their inputs, and can easily store the values for those variables at every step of the trajectory in order to condition on them so that there are no unobserved variables that could act as a confounder. 
 


\section{Conclusions and Future Work}
All three experiments have shown that the proposed intrinsic social influence reward consistently leads to higher collective return. Despite variation in the tasks, hyper-parameters, neural network architectures and experimental setups, the learning curves for agents trained with the influence reward are significantly better than the curves of powerful agents such as A3C and their improved baselines. In some cases, it is clear that influence is essential to achieve any form of learning, attesting to the promise of this idea and highlighting the complexity of learning general deep neural network multi-agent policies.

Experiment I also showed that the influence reward can lead to the emergence of communication protocols. In experiment II, which included an explicit communication channel, we saw that influence improved communication. Experiment III showed that influence can be computed by augmenting agents with an internal model of other agents. The influence reward can thus be computed without having access to another agent's reward function, or requiring a centralized controller. We were able to surpass state-of-the-art performance on the SSDs studied here, despite the fact that previous work relied on agents' ability to view other agents' rewards.


Using counterfactuals to allow agents to understand the effects of their actions on others is a promising approach with many extensions. 
Agents could use counterfactuals to develop a form of `empathy', by simulating how their actions affect another agent's value function. Influence could also be used to drive coordinated behavior in robots attempting to do cooperative manipulation and control tasks. Finally, if we view multi-agent networks as single agents, influence could be used as a regularizer to encourage different modules of the network to integrate information from other networks; for example, to hopefully prevent collapse in hierarchical RL.

\section*{Acknowledgements}
We are grateful to Eugene Vinitsky for his help in reproducing the SSD environments in open source to improve the replicability of the paper. We also thank Steven Wheelwright, Neil Rabinowitz, Thore Graepel, Alexander Novikov, Scott Reed, Pedro Mediano, Jane Wang, Max Kleiman-Weiner, Andrea Tacchetti, Kevin McKee, Yannick Schroecker, Matthias Bauer, David Rolnick, Francis Song, David Budden, and Csaba Szepesvari, as well as everyone on the DeepMind Machine Learning and Multi-Agent teams for their helpful discussions and support.

\bibliography{influence}
\bibliographystyle{icml2019}

\section{Supplementary Material}
\subsection{Influence as Mutual Information}
\label{sec:mi_connection}
The causal influence of agent $k$ on agent $j$ is:
\begin{align}
\vspace{-0.1cm}
D_{KL}\Bigl[p(a^j_{t} \mid a^k_t, z_t) \Bigl\| p(a^j_{t} \mid z_t) \Bigr] \, ,\label{eq:ci_reward}
\end{align}
where $z_t$ represents all relevant $u$ and $s$ background variables at timestep $t$. 
The influence reward to the mutual information (MI) between the actions of agents $k$ and $j$, which is given by
\begin{align}
  I(A^j;A^k|z) 
    & = \sum_{a^k, a^j} p(a^j,a^k|z) \log 
      \frac{p(a^j,a^k|z)}{p(a^j|z)p(a^k|z)} \nonumber\\
    & = \sum_{a^{k}} p(a^{k}|z)
      D_{\text{KL}} \Bigl[p(a^{j}|a^{k},z) \Bigl\| p(a^{j}|z)\Bigr],
      \label{eq:info} \, 
\end{align}
where we see that the $D_{KL}$ factor in Eq.~\ref{eq:info} is the causal influence reward given in Eq. \ref{eq:ci_reward}. 

By sampling $N$ independent trajectories $\tau_n $ from the environment, where $k$'s actions $a_n^k$ are drawn according to $p(a^k|z)$, we perform a Monte-Carlo approximation of the MI (see e.g.\ \citet{strouse2018learning}),
\begin{align}
  I(A^{k};A^{j}|z) 
  &= \mathbb{E}_{\tau}\Bigl[
    D_{\text{KL}}\bigl[ p(A^{j}|A^{k},z) \bigl\| p(A^{j}|z) \bigr] \Bigr|z \Bigr] \nonumber \\
  &\approx \frac{1}{N} \sum_n D_{\text{KL}}\bigl[ p(A^{j}|a_n^{k},z) \bigl\| p(A^{j}|z) \bigr] \, .
\label{eq:MC2}
\end{align}
Thus, in expectation, the social influence reward is the MI between agents' actions. 

Whether the policy trained with Eq.~\ref{eq:ci_reward} actually learns to approximate the MI depends on the learning dynamics. We calculate the intrinsic social influence reward using Eq.~\ref{eq:ci_reward}, because unlike Eq.~\ref{eq:info}, which gives an estimate of the symmetric bandwidth between $k$ and $j$, Eq.~\ref{eq:ci_reward} gives the directed causal effect of the specific action taken by agent $k$, $a^k_t$. We believe this will result in an easier reward to learn, since it allows for better credit assignment; agent $k$ can more easily learn which of its actions lead to high influence. 

The connection to mutual information is interesting, because a frequently used intrinsic motivation for single agent RL is \textit{empowerment}, which rewards the agent for having high mutual information between its actions and the future state of the environment (e.g.\ \citet{klyubin2005empowerment, capdepuy2007maximization}). To the extent that the social influence reward approximates the MI, $k$ is rewarded for having empowerment over $j$'s actions. 

The social influence reward can also be computed using other divergence measures besides KL-divergence. \citet{lizier2010differentiating} propose \textit{local information flow} as a measure of direct causal effect; this is equivalent to the \textit{pointwise mutual information} (the innermost term of Eq. \ref{eq:MC2}), given by:
\begin{align}
pmi(a^{k};a^{j} \mid Z=z) & =\log\frac{p(a^{j}\mid a^{k},z)}{p(a^{j} \mid z)} \nonumber \\
                          &= \log \frac{p(a^k, a^j \mid z)}{p(a^k \mid z)p(a^j \mid z)}.
\label{eq:pmi}
\end{align}
The PMI gives us a measure of influence of a single action of $k$ on the single action taken by $j$. The expectation of the PMI over $p(a^j, a^k | z)$ is the MI. We experiment with using the PMI and a number of divergence measures, including the Jensen-Shannon Divergence (JSD), and find that the influence reward is robust to the choice of measure.

\subsection{Sequential Social Dilemmas}
\label{app:ssds}

\begin{figure*}[h]
\centering
\begin{minipage}[t]{.39\textwidth}
  \centering
  \includegraphics[width=\linewidth]{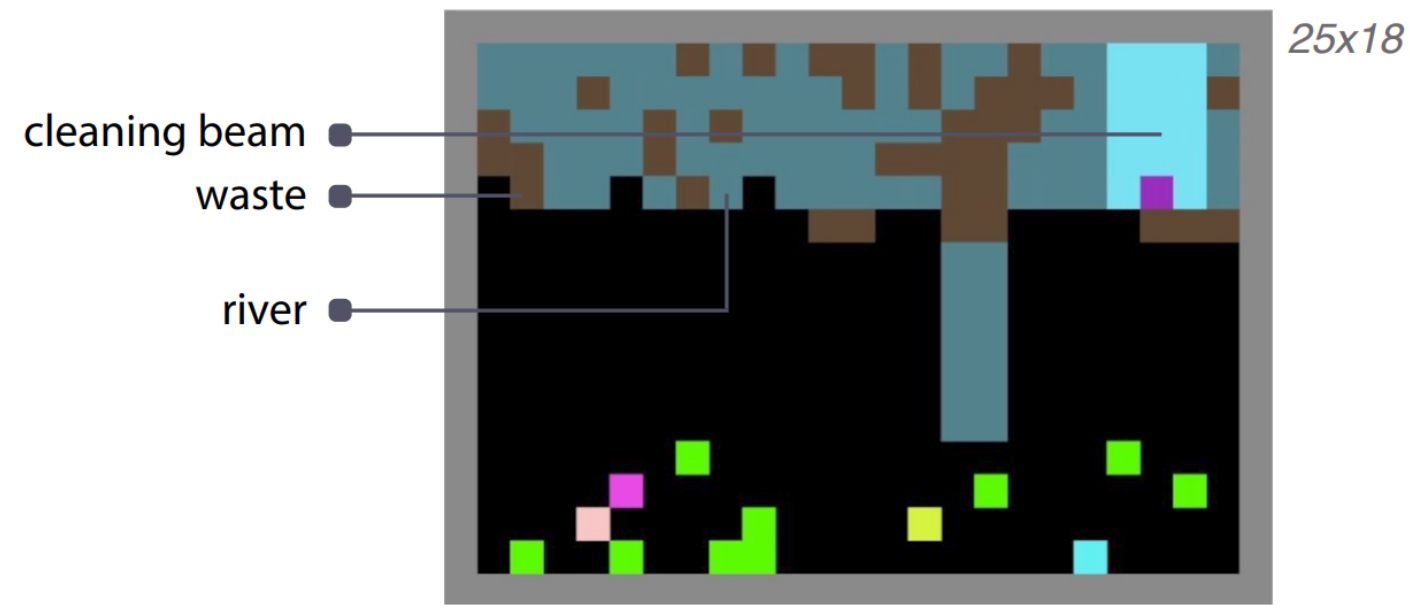}
\end{minipage}%
\begin{minipage}[t]{.59\textwidth}
  \centering
  \includegraphics[width=\linewidth]{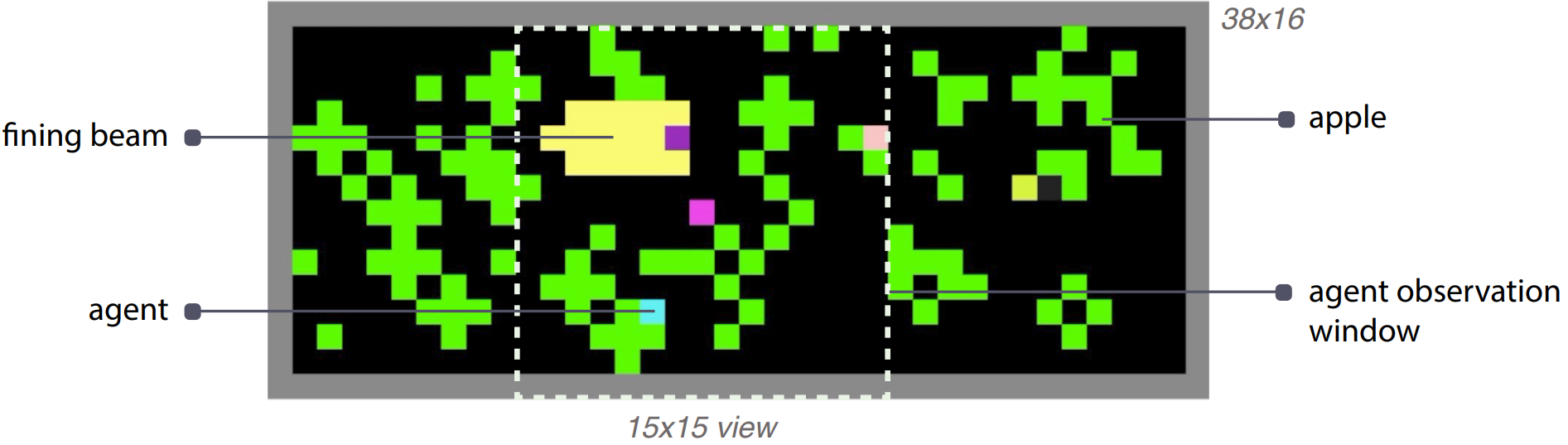}
\end{minipage}
\caption{The two SSD environments, \textit{Cleanup} (left) and \textit{Harvest} (right). Agents can exploit other agents for immediate payoff, but at the expense of the long-term collective reward of the group. Reproduced with permission from \citet{hughes2018inequity}.}
\label{fig:env}
\vspace{-0.3cm}
\end{figure*}

\begin{figure*}[h]
\vspace{-0.2cm}
\centering
\begin{subfigure}{.35\linewidth}
  \centering
  \includegraphics[width=\linewidth]{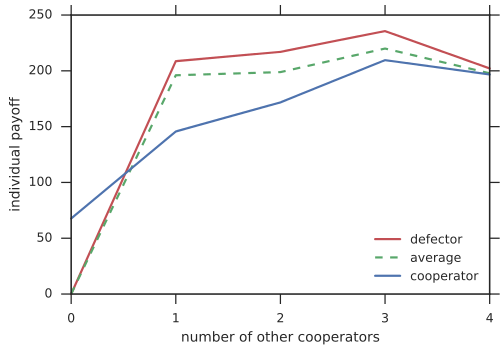}
  \vspace{-1.8cm}
  \caption{\textit{Cleanup}}
  \label{fig:schelling_huangpu}
\end{subfigure}%
\begin{subfigure}{.35\linewidth}
  \centering
  \includegraphics[width=\linewidth]{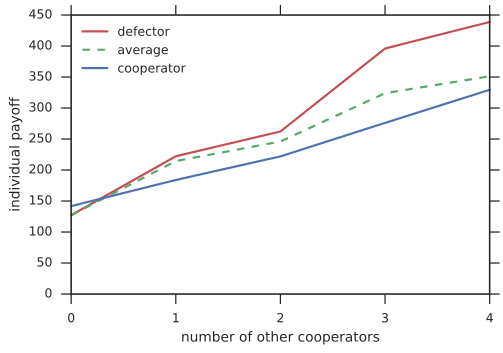}
  \vspace{-1.8cm}
  \caption{\textit{Harvest}}
  \label{fig:schelling_tragedy}
\end{subfigure}
  \vspace{8mm}
\caption{Schelling diagrams for the two social dilemma tasks show that an individual agent is motivated to defect, though everyone benefits when more agents cooperate. Reproduced 
with permission from \citet{hughes2018inequity}.}
\label{fig:schelling}
\vspace{-0.3cm}
\end{figure*}

Figure \ref{fig:env} depicts the SSD games under investigation. 
In each of the games, an agent is rewarded $+1$ for every apple it collects, but the apples are a limited resource. Agents have the ability to punish each other with a \textit{fining beam}, which costs $-1$ reward to fire, and fines any agent it hits $-50$ reward.

In \textit{Cleanup} (a public goods game) agents must clean a river before apples can grow, but are not able to harvest apples while cleaning. In \textit{Harvest} (a common pool resource game), apples respawn at a rate proportional to the amount of nearby apples; if apples are harvested too quickly, they will not grow back. Both coordination, and cooperation are required to solve both games. In \textit{Cleanup}, agents must efficiently time harvesting apples and cleaning the river, and allow agents cleaning the river a chance to consume apples. In \textit{Harvest}, agents must spatially distribute their harvesting, and abstain from consuming apples too quickly in order to harvest sustainably. 
The code for these games, including hyperparameter settings and apple and waste respawn probabilities, can be found at \url{https://github.com/eugenevinitsky/sequential\_social\_dilemma\_games}.


The reward structure of the games is shown in Figure \ref{fig:schelling}, which gives the Schelling diagram for both SSD tasks under investigation. A Schelling diagram \citep{schelling1973hockey, perolat2017multi} depicts the relative payoffs for a single agent's strategy given a fixed number of other agents who are cooperative. These diagrams show that all agents would benefit from learning to cooperate, because even the agents that are being exploited get higher reward than in the regime where all agents defect. However, traditional RL agents struggle to learn to cooperate and solve these tasks effectively \citep{hughes2018inequity}. 

\subsection{Additional experiment - Box Trapped}
\label{sec:boxtrapped}
\begin{figure}[h]
\centering
\includegraphics[width=.7\linewidth]{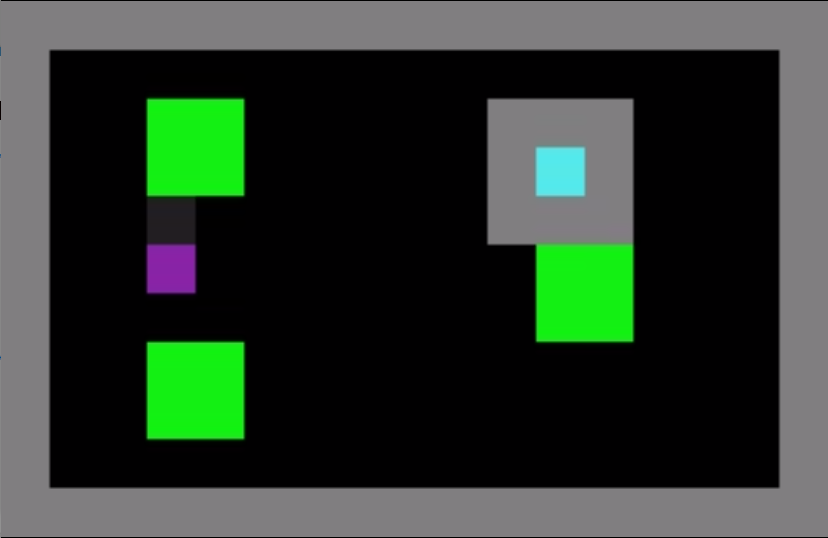} 
\caption{The \textit{Box trapped} environment in which the teal agent is trapped, and the purple agent can release it with a special \textit{open box} action.}
\label{fig:boxtrapped_env}
\end{figure}

As a proof-of-concept experiment to test whether the influence reward works as expected, we constructed a special environment, shown in Figure \ref{fig:boxtrapped_env}. In this environment, one agent (teal) is trapped in a box. The other agent (purple) has a special action it can use to open the box... or it can simply choose to consume apples, which exist outside the box and are inexhaustible in this environment. 

As expected, a vanilla A3C agent learns to act selfishly; the purple agent will simply consume apples, and  chooses the \textit{open box} action in 0\% of trajectories once the policy has converged. A video of A3C agents trained in this environment is available at: \url{https://youtu.be/C8SE9_YKzxI}, which shows that the purple agent leaves its compatriot trapped in the box throughout the trajectory.

\begin{figure}[h]
\centering
  \includegraphics[width=0.9\linewidth]{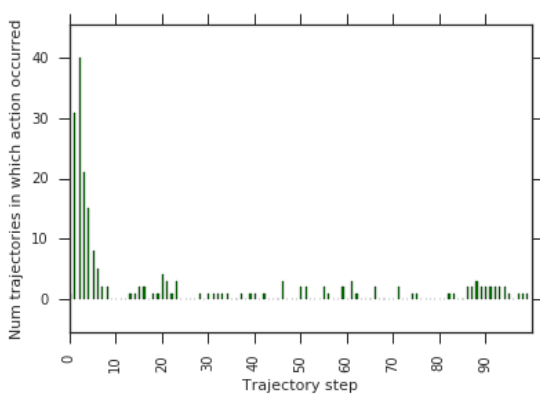}
  \caption{Number of times the \textit{open box} action occurs at each trajectory step over 100 trajectories.}
  \label{fig:open_box_step}
\end{figure}

In contrast, an agent trained with the social influence reward chooses the \textit{open box} action in 88\% of trajectories, releasing its fellow agent so that they are both able to consume apples. A video of this behavior is shown at: \url{https://youtu.be/Gfo248-qt3c}. Further, as Figure \ref{fig:open_box_step} reveals, the purple influencer agent usually chooses to open the box within the first few steps of the trajetory, giving its fellow agent more time to collect reward. 

\begin{figure}[h]
  \centering
  \includegraphics[width=0.9\linewidth]{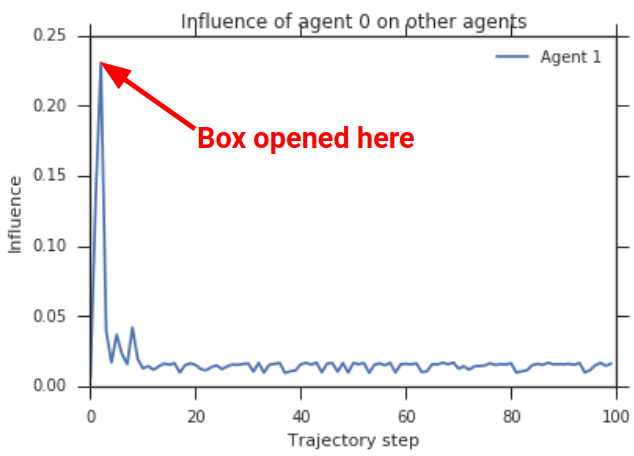}
  \caption{Influence reward over a trajectory in \textit{Box trapped}. An agent gets high influence for letting another agent out of the box in which it is trapped.}
  \label{fig:boxtrapped_influence}
\end{figure}

Most importantly though, Figure \ref{fig:boxtrapped_influence} shows the influence reward over the course of a trajectory in the \textit{Box trapped} environment. The agent chooses the \textit{open box} action in the second timestep; at this point, we see a corresponding spike in the influence reward. This reveals that the influence reward works as expected, incentivizing an action which has a strong --- and in this case, prosocial --- effect on the other agent's behavior.

\subsection{Implementation details} 
\label{app:implementation}
All models are trained with a single convolutional layer with a kernel of size 3, stride of size 1, and 6 output channels. This is connected to two fully connected layers of size 32 each, and an LSTM with 128 cells. We use a discount factor $\gamma=.99$. The number of agents $N$ is fixed to 5. 

In addition to the comparison function used to compute influence (e.g. KL-divergence, PMI, JSD), there are many other hyperparameters that can be tuned for each model. We use a random search over hyperparameters, ensuring a fair comparison with the search size over the baseline parameters that are shared with the influence models. For all models we search for the optimal entropy reward and learning rate, where we anneal the learning rate from an initial value \texttt{lr\_init} to \texttt{lr\_final}. The below sections give the parameters found to be most effective for each of the three experiments. 

\subsubsection{Basic influence hyperparameters}
\label{sec:cc_hparams}
In this setting we vary the number of influencers from $1-4$, the influence reward weight $\beta$, and the number of curriculum steps over which the weight of the influence reward is linearly increased $C$. In this setting, since we have a centralised controller, we also experiment with giving the influence reward to the agent being influenced as well, and find that this sometimes helps. This `influencee' reward is not used in the other two experiments, since it precludes independent training. The hyperparameters found to give the best performance for each model are shown in Table \ref{tab:hparams_cc}.

\begin{table*}[h]
\centering
\small
\begin{tabular}{l|lll|lll|}
                         & \multicolumn{3}{c}{Cleanup}                                                                                                              & \multicolumn{3}{c|}{Harvest}                                                                                                              \\ \cline{2-7} 
Hyperparameter           & \begin{tabular}[c]{@{}l@{}}A3C \\ baseline\end{tabular} & \begin{tabular}[c]{@{}l@{}}Visible actions\\ baseline\end{tabular} & Influence & \begin{tabular}[c]{@{}l@{}}A3C \\ baseline\end{tabular} & \begin{tabular}[c]{@{}l@{}}Visible actions \\ baseline\end{tabular} & Influence \\ \hline
Entropy reg.             & .00176                                                  & .00176                                                             & .000248   & .000687                                                 & .00184                                                              & .00025    \\
lr\_init                 & .00126                                                  & .00126                                                             & .00107    & .00136                                                  & .00215                                                              & .00107    \\
lr\_end                  & .000012                                                 & .000012                                                            & .000042   & .000028                                                 & .000013                                                             & .000042   \\
Number of influencers    & -                                                       & 3                                                                  & 1         & -                                                       & 3                                                                   & 3         \\
Influence weight $\beta$ & -                                                       & 0                                                                  & .146      & -                                                       & 0                                                                   & .224      \\
Curriculum $C$           & -                                                       & -                                                                  & 140       & -                                                       & -                                                                   & 140       \\
Policy comparison        & -                                                       & -                                                                  & JSD       & -                                                       & -                                                                   & PMI       \\
Influencee reward        & -                                                       & \textbf{-}                                                         & 1         & -                                                       & -                                                                   & 0        
\end{tabular}
\caption{Optimal hyperparameter settings for the models in the basic influence experiment.}
\label{tab:hparams_cc}
\end{table*}

\subsubsection{Communication hyperparameters}
Because the communication models have an extra A2C output head for the communication policy, we use an additional entropy regularization term just for this head, and apply a weight to the communication loss in the loss function. We also vary the number of communication symbols that the agents can emit, and the size of the linear layer that connects the LSTM to the communication policy layer, which we term the communication embedding size. Finally, in the communication regime, we experiment to setting the weight on the extrinsic reward E, $\alpha$, to zero. The best hyperparameters for each of the communication models are shown in Table \ref{tab:hparams_comm}.

\begin{table*}[h]
\centering
\small
\begin{tabular}{l|lll|lll|}
                                                                            & \multicolumn{3}{c}{Cleanup}                                                                                                                                                    & \multicolumn{3}{c|}{Harvest}                                                                                                                                                   \\ \cline{2-7} 
Hyperparameter                                                              & \begin{tabular}[c]{@{}l@{}}A3C \\ baseline\end{tabular} & \begin{tabular}[c]{@{}l@{}}Comm.\\ baseline\end{tabular} & \begin{tabular}[c]{@{}l@{}}Influence\\ comm.\end{tabular} & \begin{tabular}[c]{@{}l@{}}A3C \\ baseline\end{tabular} & \begin{tabular}[c]{@{}l@{}}Comm.\\ baseline\end{tabular} & \begin{tabular}[c]{@{}l@{}}Influence\\ comm.\end{tabular} \\ \hline
Entropy reg.                                                                & .00176                                                  & .000249                                                  & .00305                                                    & .000687                                                 & .000174                                                  & .00220                                                    \\
lr\_init                                                                    & .00126                                                  & .00223                                                   & .00249                                                    & .00136                                                  & .00137                                                   & .000413                                                   \\
lr\_end                                                                     & .000012                                                 & .000022                                                  & .0000127                                                  & .000028                                                 & .0000127                                                 & .000049                                                   \\
Influence weight $\beta$                                                    & -                                                       & 0                                                        & 2.752                                                     & -                                                       & 0                                                        & 4.825                                                     \\
\begin{tabular}[c]{@{}l@{}}Extrinsic reward \\ weight $\alpha$\end{tabular} & -                                                       & -                                                        & 0                                                         & -                                                       & -                                                        & 1.0                                                       \\
Curriculum $C$                                                              & -                                                       & -                                                        & 1                                                         & -                                                       & -                                                        & 8                                                         \\
Policy comparison                                                           & -                                                       & -                                                        & KL                                                        & -                                                       & -                                                        & KL                                                        \\
Comm. entropy reg.                                                          & -                                                       & \textbf{-}                                               & .000789                                                   & -                                                       & -                                                        & .00208                                                    \\
Comm. loss weight                                                           & -                                                       & -                                                        & .0758                                                     & -                                                       & -                                                        & .0709                                                     \\
Symbol vocab size                                                           & -                                                       & -                                                        & 9                                                         & -                                                       & -                                                        & 7                                                         \\
Comm. embedding                                                             & -                                                       & -                                                        & 32                                                        & -                                                       & -                                                        & 16                                                       
\end{tabular}
\caption{Optimal hyperparameter settings for the models in the communication experiment.}
\label{tab:hparams_comm}
\end{table*}

\subsubsection{Model of other agents (MOA) hyperparameters}
The MOA hyperparameters include whether to only train the MOA with cross-entropy loss on the actions of agents that are visible, and how much to weight the supervised loss in the overall loss of the model. The best hyperparameters are shown in Table \ref{tab:hparams_moa}.

\begin{table*}[h]
\centering
\small
\begin{tabular}{l|lll|lll|}
                                                                       & \multicolumn{3}{c}{Cleanup}                                                                                                                                                & \multicolumn{3}{c|}{Harvest}                                                                                                                                               \\ \cline{2-7} 
Hyperparameter                                                         & \begin{tabular}[c]{@{}l@{}}A3C \\ baseline\end{tabular} & \begin{tabular}[c]{@{}l@{}}MOA\\ baseline\end{tabular} & \begin{tabular}[c]{@{}l@{}}Influence\\ MOA\end{tabular} & \begin{tabular}[c]{@{}l@{}}A3C \\ baseline\end{tabular} & \begin{tabular}[c]{@{}l@{}}MOA\\ baseline\end{tabular} & \begin{tabular}[c]{@{}l@{}}Influence\\ MOA\end{tabular} \\ \hline
Entropy reg.                                                           & .00176                                                  & .00176                                                 & .00176                                                  & .000687                                                 & .00495                                                 & .00223                                                  \\
lr\_init                                                               & .00126                                                  & .00123                                                 & .00123                                                  & .00136                                                  & .00206                                                 & .00120                                                  \\
lr\_end                                                                & .000012                                                 & .000012                                                & .000012                                                 & .000028                                                 & .000022                                                & .000044                                                 \\
Influence weight $\beta$                                               & -                                                       & 0                                                      & .620                                                    & -                                                       & 0                                                      & 2.521                                                   \\
MOA loss weight                                                        & -                                                       & 1.312                                                  & 15.007                                                  & -                                                       & 1.711                                                  & 10.911                                                  \\
Curriculum $C$                                                         & -                                                       & -                                                      & 40                                                      & -                                                       & -                                                      & 226                                                     \\
Policy comparison                                                      & -                                                       & -                                                      & KL                                                      & -                                                       & -                                                      & KL                                                      \\
\begin{tabular}[c]{@{}l@{}}Train MOA only \\ when visible\end{tabular} & -                                                       & False                                                  & True                                                    & -                                                       & False                                                  & True                                                   
\end{tabular}
\caption{Optimal hyperparameter settings for the models in the model of other agents (MOA) experiment.}
\label{tab:hparams_moa}
\end{table*}

\subsubsection{Communication analysis}
\label{app:comm}
The speaker consistency metric is calculated as: 
\begin{align}
    \sum_{k=1}^N 0.5[&\sum_{c} 1- \frac{H(p(a^k|m^k=c))}{H_{max}} \nonumber \\
    &+ \sum_a 1- \frac{H(p(m^k|a^k=a))}{H_{max}}],
\end{align} 
where $H$ is the entropy function and $H_{max}$ is the maximum entropy based on the number of discrete symbols or actions. The goal of the metric is to measure how much of a 1:1 correspondence exists between a speaker's action and the speaker's communication message. 

\subsection{Additional results}
\label{app:add_results}

\subsubsection{Basic influence emergent communication}

\begin{figure}[h]
\begin{centering}
\includegraphics[width=.5\linewidth]{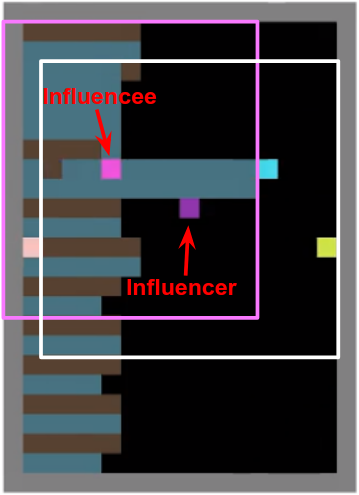}
\caption{A moment of high influence between the purple influencer and magenta influencee.}
\label{fig:influence_moment2}
\end{centering}
\end{figure}

Figure \ref{fig:influence_moment2} shows an additional moment of high influence in the \textit{Cleanup} game. The purple influencer agent can see the area within the white box, and therefore all of the apple patch. The field-of-view of the magenta influencee is outlined with the magenta box; it cannot see if apples have appeared, even though it has been cleaning the river, which is the action required to cause apples to appear. When the purple influencer turns left and does not move towards the apple patch, this signals to the magenta agent that no apples have appeared, since otherwise the influence would move right. 

\subsubsection{Optimizing for collective reward}

\begin{figure}[h]
\centering
\begin{subfigure}{.5\linewidth}
  \centering
  \includegraphics[width=\linewidth]{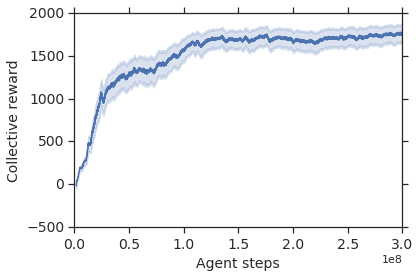}
  \caption{\textit{Cleanup}}
  \label{fig:group_cleanup}
\end{subfigure}%
\begin{subfigure}{.5\linewidth}
  \centering
  \includegraphics[width=\linewidth]{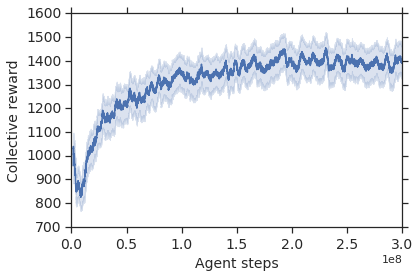}
  \caption{\textit{Tragedy}}
  \label{fig:group_harvest}
\end{subfigure}
\caption{Total collective reward obtained by agents trained to optimize for the collective reward, for the 5 best hyperparameter settings with 5 random seeds each. Error bars show a 99.5\% confidence interval (CI) computed within a sliding window of 200 agent steps.}
\label{fig:group_return}
\end{figure}

In this section we include the results of training explicitly prosocial agents, which  directly optimize for the collective reward of all agents. Previous work (e.g. \citet{peysakhovich2018prosocial}) has shown that training agents to optimize for the rewards of other agents can help the group to obtain better collective outcomes. Following a similar principle, we implemented agents that optimize for a convex combination of their own individual reward $e^k_t$ and the collective reward of all other agents, $\sum_{i=1, i \neq k}^N e^i_t$. Thus, the reward function for agent $k$ is $r^k_t = e^k_t + \eta \sum_{i=1, i \neq k}^N e^i_t$. We conducted the same hyperparameter search over the parameters mentioned in Section \ref{sec:cc_hparams} varying the weight placed on the collective reward, $\eta \in [0,2]$.

As expected, we find that agents trained to optimize for collective reward attain higher collective reward in both \textit{Cleanup} and \textit{Harvest}, as is shown in Figure \ref{fig:group_return}. In both games, the optimal value for $\eta=0.85$. Interestingly, however, the equality in the individual returns for these agents is extremely low. Across the hyperparameter sweep, no solution to the \textit{Cleanup} game which scored more than 20 points in terms of collective return was found in which all agents scored an individual return above 0. It seems that in \textit{Cleanup}, when agents are trained to optimize for collective return, they converge on a solution in which some agents never receive any reward.

Note that training agents to optimize for collective reward requires that each agent can view the rewards obtained by other agents. As discussed previously, the social influence reward is a novel way to obtain cooperative behavior, that does not require making this assumption. 

\subsubsection{Collective reward and equality}
It is important to note that collective reward is not always the perfect metric of cooperative behavior, a finding that was also discovered by \citet{barton2018measuring} and emphasized by \citet{leibo2017multi}. In the case, we find that there is a spurious solution to the \textit{Harvest} game, in which one agent fails to learn and fails to collect any apples. This leads to very high collective reward, since it means there is one fewer agent that can exploit the others, and makes sustainable harvesting easier to achieve. Therefore, for the results shown in the paper, we eliminate any random seed in \textit{Harvest} for which one of the agents has failed to learn to collect apples, as in previous work \citep{hughes2018inequity}. 

\begin{figure*}[h] 
\begin{subfigure}{0.48\textwidth}
\includegraphics[width=\linewidth]{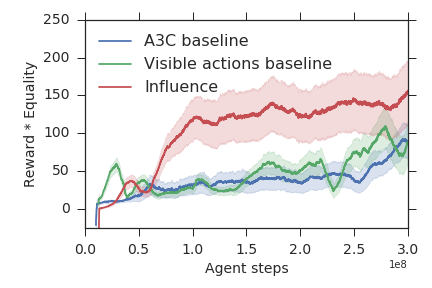}
\caption{\textit{Cleanup} - Basic influence} \label{fig:app_results_cc_huangpu}
\end{subfigure}\hspace*{\fill}
\begin{subfigure}{0.48\textwidth}
\includegraphics[width=\linewidth]{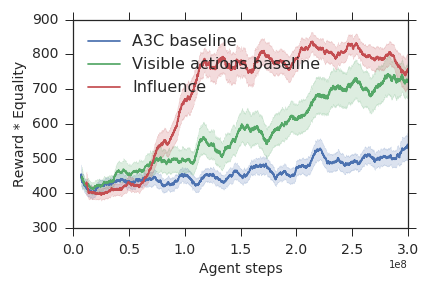}
\caption{\textit{Harvest} - Basic influence} \label{fig:app_results_cc_huangpu}
\end{subfigure}

\medskip
\begin{subfigure}{0.48\textwidth}
\includegraphics[width=\linewidth]{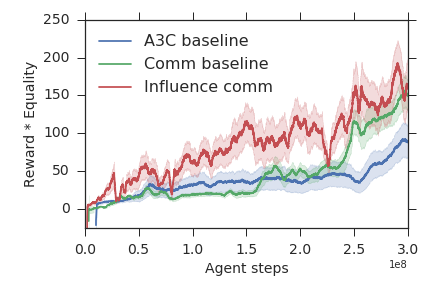}
\caption{\textit{Cleanup} - Communication} \label{fig:app_results_comm_huangpu}
\end{subfigure}\hspace*{\fill}
\begin{subfigure}{0.48\textwidth}
\includegraphics[width=\linewidth]{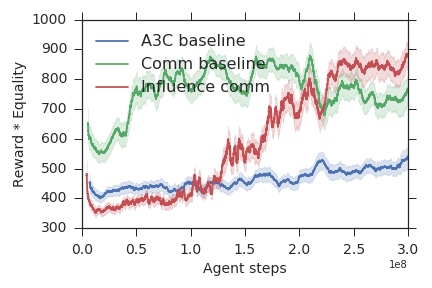}
\caption{\textit{Harvest} - Communication} \label{fig:app_results_comm_huangpu}
\end{subfigure}

\medskip
\begin{subfigure}{0.48\textwidth}
\includegraphics[width=\linewidth]{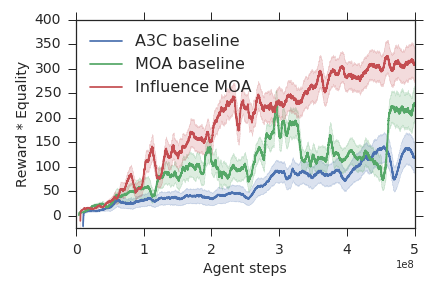}
\caption{\textit{Cleanup} - Model of other agents} \label{fig:app_results_tom_huangpu}
\end{subfigure}\hspace*{\fill}
\begin{subfigure}{0.48\textwidth}
\includegraphics[width=\linewidth]{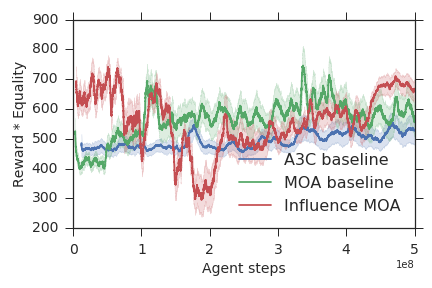}
\caption{\textit{Harvest} - Model of other agents} \label{fig:app_results_tom_huangpu}
    \end{subfigure}

\caption{Total collective reward times equality, $R * (1-G)$, obtained in all experiments. Error bars show a 99.5\% confidence interval (CI) over 5 random seeds, computed within a sliding window of 200 agent steps. Once again, the models trained with influence reward (red) significantly outperform the baseline and ablated models.} 
\label{fig:app_all_results_rxe}
\end{figure*}

However, here we also present an alternative strategy for assessing the overall collective outcomes: weighting the total collective reward by an index of equality of the individual rewards. Specifically, we compute the Gini coefficient over the $N$ agents' individual environmental rewards $e^k_t$:
\begin{align}
    G = \frac{\sum_{i=1}^N \sum_{j=1}^N |e^i_t - e^j_t|}{2N\sum_{i=1}^N e^i_t},
\end{align}
which gives us a measure of the inequality of the returns, where $G \in [0,1]$, with $G=0$ indicating perfect equality. Thus, $1-G$ is a measure of equality; we use this to weight the collective reward for each experiment, and plot the results in Figure \ref{fig:app_all_results_rxe}. Once again, we see that the influence models give the highest final performance, even with this new metric. 

\subsubsection{Collective reward over multiple hyperparameters}
\begin{figure*}[h] 
\begin{subfigure}{0.48\textwidth}
\includegraphics[width=\linewidth]{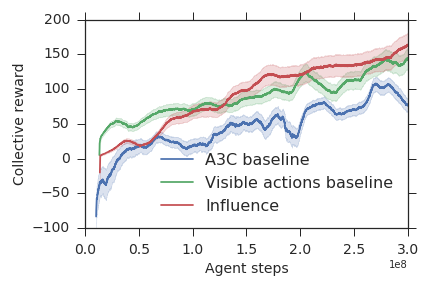}
\caption{\textit{Cleanup} - Basic influence} \label{fig:app_results_cr5_cc_huangpu}
\end{subfigure}\hspace*{\fill}
\begin{subfigure}{0.48\textwidth}
\includegraphics[width=\linewidth]{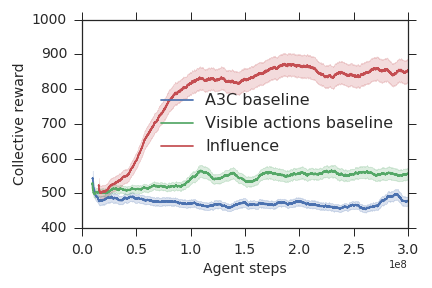}
\caption{\textit{Harvest} - Basic influence} \label{fig:app_results_cr5_cc_huangpu}
\end{subfigure}

\medskip
\begin{subfigure}{0.48\textwidth}
\includegraphics[width=\linewidth]{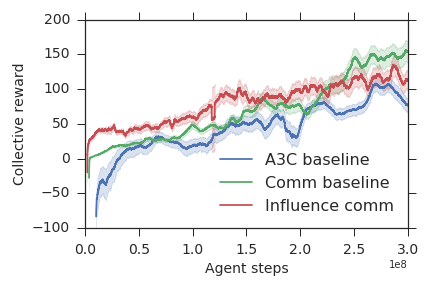}
\caption{\textit{Cleanup} - Communication} \label{fig:app_results_cr5_comm_huangpu}
\end{subfigure}\hspace*{\fill}
\begin{subfigure}{0.48\textwidth}
\includegraphics[width=\linewidth]{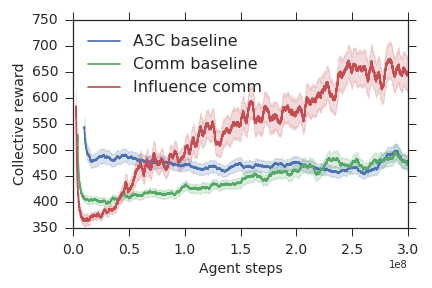}
\caption{\textit{Harvest} - Communication} \label{fig:app_results_cr5_comm_huangpu}
\end{subfigure}

\medskip
\begin{subfigure}{0.48\textwidth}
\includegraphics[width=\linewidth]{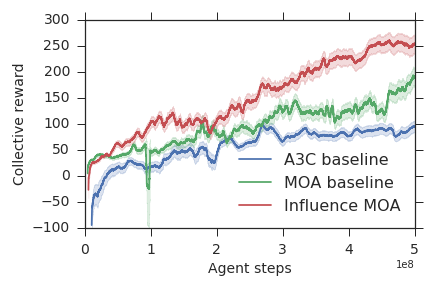}
\caption{\textit{Cleanup} - Model of other agents} \label{fig:app_results_cr5_tom_huangpu}
\end{subfigure}\hspace*{\fill}
\begin{subfigure}{0.48\textwidth}
\includegraphics[width=\linewidth]{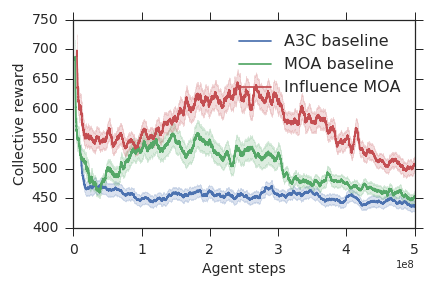}
\caption{\textit{Harvest} - Model of other agents} \label{fig:app_results_cr5_tom_huangpu}
    \end{subfigure}

\caption{Total collective reward over the top 5 hyperparameter settings, with 5 random seeds each, for all experiments. Error bars show a 99.5\% confidence interval (CI) computed within a sliding window of 200 agent steps. The influence models still maintain an advantage over the baselines and ablated models, suggesting the technique is robust to the hyperparameter settings.} 
\label{fig:app_all_results_cr5}
\end{figure*}

Finally, we would like to show that the influence reward is robust to the choice of hyperparameter settings. Therefore, in Figure \ref{fig:app_all_results_cr5}, we plot the collective reward of the top 5 best hyperparameter settings for each experiment, over 5 random seeds each. Once again, the influence models result in higher collective reward, which provides evidence that the model is robust to the choice of hyperparameters. 

\subsubsection{Performance comparison between models and related work}
Table \ref{tab:final_scores} presents the final collective reward obtained by each of the models tested in the experiments presented in the paper. We see that in several cases, the influence agents are even able to out-perform the state-of-the-art results on these tasks reported by \cite{hughes2018inequity}, despite the fact that the solution proposed by \cite{hughes2018inequity} requires that agents can view other agents' rewards, whereas we do not make this assumption, and instead only require that agents can view each others' actions. 

\begin{table}[h]
\centering
\begin{tabular}{lll}
                           & Cleanup        & Harvest       \\ \hline
A3C baseline               & 89           & 485           \\
Inequity aversion (Hughes et al.)         & 275          & 750           \\
Influence - Basic          & 190          & \textbf{1073} \\
Influence - Communication  & 166          & \textbf{951}           \\
Influence - Model of other agents & \textbf{392} & 588    
\end{tabular}
\caption{Final collective reward over the last 50 agent steps for each of the models considered. Bolded entries represent experiments in which the influence models significantly outperformed the scores reported in previous work on \textit{inequity aversion} \citep{hughes2018inequity}. This is impressive, considering the \textit{inequity averse} agents are able to view all other agents' rewards. We make no such assumption, and yet are able to achieve similar or superior performance.}
\label{tab:final_scores}
\end{table}

\end{document}